\documentclass[pdflatex,sn-mathphys-num]{sn-jnl}

\usepackage{graphicx}%
\usepackage{multirow}%
\usepackage{amsmath,amssymb,amsfonts}%
\usepackage{amsthm}%
\usepackage{mathrsfs}%
\usepackage[title]{appendix}%
\usepackage{xcolor}%
\usepackage{textcomp}%
\usepackage{manyfoot}%
\usepackage{booktabs}%
\usepackage{algorithm}%
\usepackage{algorithmicx}%
\usepackage{algpseudocode}%
\usepackage{listings}%
\usepackage{comment}%
\usepackage[export]{adjustbox}

\theoremstyle{thmstyleone}%
\theoremstyle{thmstyletwo}%

\theoremstyle{thmstylethree}%

\usepackage{xcolor}
\usepackage[normalem]{ulem}

\begin{document}

\title[Article Title]{AI-Powered Prediction of Nanoparticle Pharmacokinetics: A Multi-View Learning Approach}


\author[1]{\fnm{Amirhossein Khakpour}}\email{ak01905@surrey.ac.uk}

\author[1]{\fnm{Lucia Florescu}}\email{l.m.florescu@surrey.ac.uk}

\author[4]{\fnm{Richard Tilley}}\email{r.tilley@unsw.edu.au}

\author[2]{\fnm{Haibo Jiang}}\email{hbjiang@hku.hk}

\author*[3]{\fnm{K. Swaminathan Iyer}}\email{swaminatha.iyer@uwa.edu.au}

\author*[1]{\fnm{Gustavo Carneiro}}\email{g.carneiro@surrey.ac.uk}

\abstract{
 
The clinical translation of nanoparticle-based treatments remains limited due to the unpredictability of (nanoparticle) NP pharmacokinetics—how they distribute, accumulate, and clear from the body. Predicting these behaviours is challenging due to complex biological interactions and the difficulty of obtaining high-quality experimental datasets.
Existing AI-driven approaches rely heavily on data-driven learning but fail to integrate crucial knowledge about NP properties and biodistribution mechanisms. We introduce a multi-view deep learning framework that enhances pharmacokinetic predictions by incorporating prior knowledge of key NP properties such as size and charge  into a cross-attention mechanism, enabling context-aware feature selection and improving generalization despite small datasets. To further enhance prediction robustness, we employ an ensemble learning approach, combining deep learning with XGBoost (XGB) and Random Forest (RF), which significantly outperforms existing AI models. Our interpretability analysis reveals key physicochemical properties driving NP biodistribution, providing biologically meaningful insights into possible mechanisms governing NP behaviour in vivo rather than a black-box model. Furthermore, by bridging machine learning with physiologically based pharmacokinetic (PBPK) modelling, this work lays the foundation for data-efficient AI-driven drug discovery and precision nanomedicine.}




\maketitle

\section{Introduction}\label{sec:introduction}

NP-based drug delivery systems have revolutionized modern medicine, offering highly targeted therapies with reduced side effects~\cite{peer2020nanocarriers, brigger2012nanoparticles, gavas2021nanoparticles}. NP imaging agents, ranging from magnetic NP contrast agents to light-emitting quantum dots, have the potential to transform how we medically diagnose and monitor serious diseases within the body~\cite{wang2010targeting, sztandera2018gold}. However, the clinical translation of NP-based treatments and imaging remains limited due to the unpredictability of NP pharmacokinetics—how NPs distribute, accumulate, and clear from the body~\cite{davis2008nanoparticle, yuan2023pharmacokinetics}. Traditional PBPK models attempt to predict NP behavior, but they rely on predefined equations and assumptions that fail to fully capture the complexity of biological variability, tumor microenvironments, and organ interactions~\cite{chen2023meta, wang2013metabolism}. This gap in predictive accuracy has hindered the development of optimized NP formulations for drug delivery~\cite{yuan2023pharmacokinetics}.

Machine learning has emerged as a powerful tool to improve pharmacokinetic predictions by leveraging large datasets to learn complex relationships between NP properties and their biodistribution~\cite{he2024additive, chou2023artificial}. However, existing AI-based approaches face two major challenges: limited data availability, as pharmacokinetic studies involve resource-intensive in vivo experiments, and lack of prior knowledge integration, meaning that traditional ML models treat each dataset independently without leveraging fundamental insights from nanomedicine~\cite{jyakhwo2024machine, rezvantalab2024machine}. As a result, existing ML models struggle to generalize beyond specific experimental conditions, limiting their applicability in real-world clinical scenarios~\cite{chou2023artificial, serretiello2024extracellular}.

\begin{figure}[!h]
    \centering
    \includegraphics[width=\linewidth]{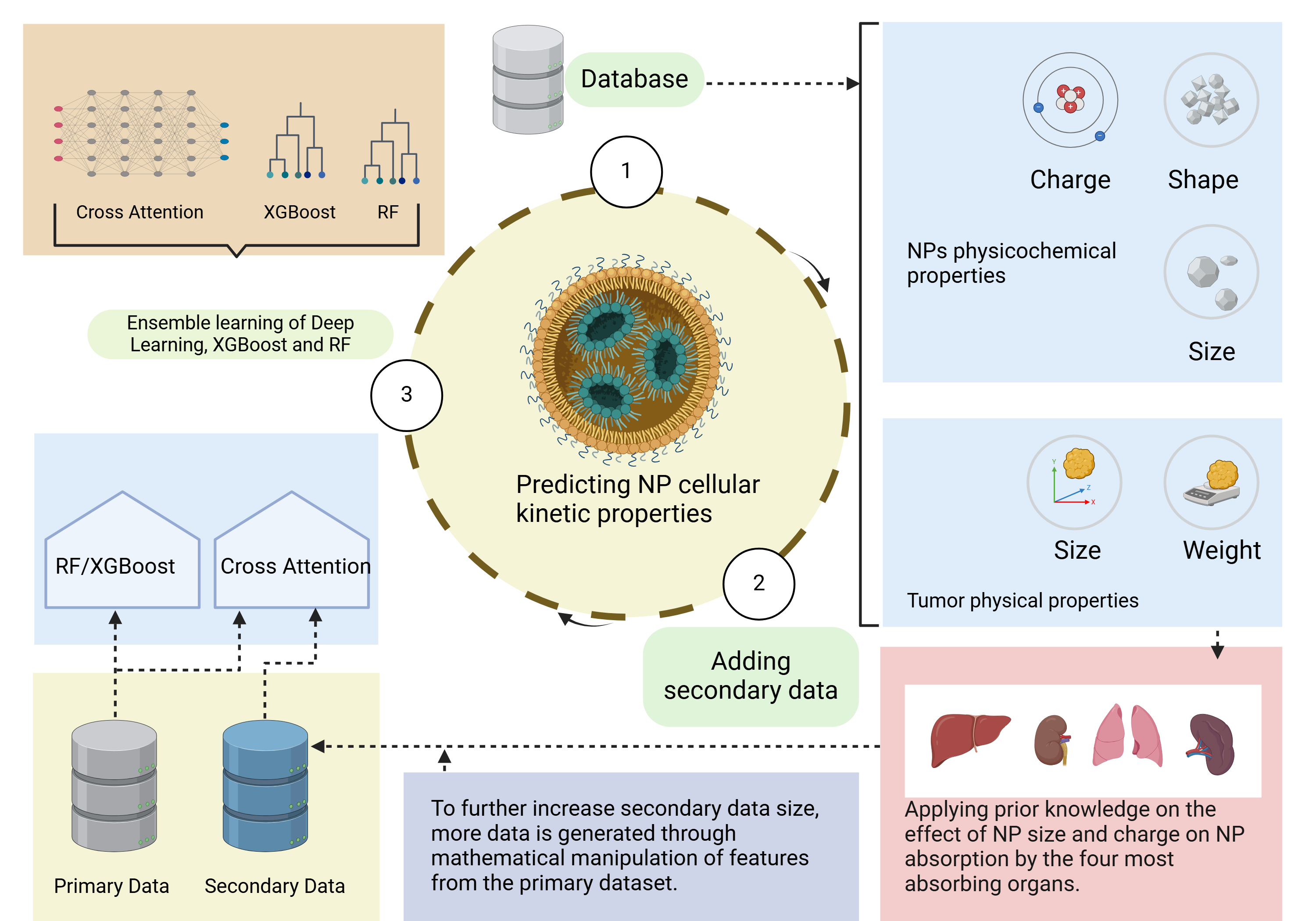}
    \caption{
    Overview of the multi-view AI framework. The primary dataset consists of NP physicochemical properties and pharmacokinetic measurements, while prior knowledge is incorporated through engineered features. A cross-attention deep learning model integrates these two data sources, and ensemble learning improves overall prediction robustness. Created in BioRender. Khakpour, A. (2025) https://BioRender.com/c18t496}
    \label{fig:introduction}
\end{figure}

To address these challenges, we introduce a multi-view AI framework that combines prior knowledge of NP properties with cross-attention deep learning to enhance pharmacokinetic predictions, even in data-limited scenarios~\cite{lin2022predicting, du2018effect}--see Fig.~\ref{fig:introduction}. This multi-view framework approach enables the model to leverage nanomedical domain-specific insights such as the importance of NP size, charge, shape, and organ biodistribution patterns to refine its predictions beyond purely data-driven learning~\cite{choi2007renal, sonavane2008biodistribution}. The cross-attention mechanism assigns importance to key biological and physicochemical properties, improving interpretability and accuracy~\cite{vaswani2017attention}. The model also integrates an ensemble learning approach, combining deep learning with classical machine learning models such as RF and XGB to enhance robustness and predictive performance~\cite{ganaie2022ensemble}.

Building on this framework, the proposed model predicts four key NP pharmacokinetic parameters—\({KTRESmax}\), \({KTRESn}\), \({KTRES50}\), and \({KTRESrelease}\), which describe NP biodistribution across organs, including the maximum uptake rate constant, Hill coefficient, time to reach half-max uptake, and release rate constant in tumour cells~\cite{chou2023artificial}. By linking machine learning with PBPK modelling, this approach not only improves pharmacokinetic predictions but also lays the foundation for AI-assisted drug formulation, NP optimization, and precision nanomedicine~\cite{noorain2023machine, lin2022predicting}. This study presents the first AI-powered pharmacokinetic model that systematically integrates prior knowledge with deep learning, setting a new benchmark for NP biodistribution modelling~\cite{he2024additive, chen2023meta, lin2022predicting}.

\subsection{Background}
PBPK models simulate NP absorption, metabolism, and elimination in various organs~\cite{yuan2023pharmacokinetics, wang2013metabolism, chou2023artificial, li2017physiologically}. However, their reliance on simplified assumptions limits their accuracy in complex environments~\cite{chen2023meta, yuan2023pharmacokinetics, wang2013metabolism}. PBPK models struggle to incorporate dynamic changes in NP behaviour, such as real-time cellular uptake or targeting influenced by surface modifications~\cite{yuan2023pharmacokinetics, cheng2020meta}. These challenges have led to growing interest in AI-driven approaches~\cite{he2024additive, chou2023artificial}.

AI models have increasingly been integrated with PBPK modelling to reduce reliance on in vivo animal data~\cite{chou2023artificial}. Despite these advances, challenges remain due to variability in NP behaviour across biological environments~\cite{chen2023meta, yuan2023pharmacokinetics}. The unpredictable uptake and elimination of NPs add complexity to PBPK modelling~\cite{davis2008nanoparticle, zhang2016nanoparticle, wang2013metabolism}.

While AI models leveraging meta-analysis data have improved NP distribution predictions~\cite{chen2023meta}, meta-analysis alone does not account for critical NP properties such as size and charge, which significantly influence organ-specific biodistribution. Feature engineering improves model performance through polynomial, logarithmic, and exponential transformations~\cite{brambor2006understanding, kohli1991reservation, fama1992cross, huber1985projection, box1964analysis}. By integrating domain-specific NP properties with AI-driven approaches, models can better capture NP pharmacokinetics, addressing limitations of both PBPK and traditional machine learning methods.

\section{Results and Discussion}
\label{sec:results_and_discussion}



The proposed multi-view AI framework was evaluated against baseline models, including standard deep learning architectures and traditional ML methods such as RF and XGB, using performance metrics like mean squared error and \( R^2 \) values to assess prediction accuracy. In this section, we first describe the dataset and experimental setup, followed by details of the ensemble method. We then present the results, conduct an ablation study, and explore the impact of multi-task learning on training the four pharmacokinetic parameters.

\subsection{Dataset}

Our study utilizes a publicly available pharmacokinetic dataset consisting of 378 samples in mice~\cite{chou2023artificial}. Each sample includes NP physicochemical properties such as size, charge, and shape, along with corresponding pharmacokinetic parameters, including the maximum uptake rate constant and release rate constant~\cite{chou2023artificial}. Table~\ref{tab:data} shows details about the input features and prediction values of the dataset.

For model evaluation, the primary dataset undergoes pre-processing, where samples with missing values are removed, resulting in a final dataset of 280 samples. This dataset is then split into train-validation-test partitions, as in: 60\% for training, 20\% for validation, and 20\% for testing. 
We also run a 5-fold cross validation experiment for model selection, where hyperparameters are estimated. We show results from both the train-validation-test and 5-fold cross-validation experiments.
The mean squared error loss function is minimized using an Adam or SGD optimizer, with the learning rate selected from \{1e\textsuperscript{-3}, 5e\textsuperscript{-4}, 1e\textsuperscript{-4}\}.

To enhance predictive accuracy despite the limited dataset size, we integrate prior knowledge from domain-specific studies on NP interactions with biological systems. This prior knowledge is encoded into additional engineered features, reflecting well-documented relationships between NP physicochemical properties and their biodistribution.

To address data imbalance and improve model robustness, we employ synthetic minority over-sampling technique (SMOTE) for data augmentation, particularly for underrepresented pharmacokinetic response values~\cite{torgo2013smote}.



\begin{table}[h]
\caption{Details of the dataset input features and prediction values, which contain various categorical and numerical features. Adapted from~\cite{chou2023artificial}.
}\label{tab:data}%
\begin{tabular}{@{}p{3cm}p{2cm}p{2cm}p{5cm}@{}}
\toprule
Input Feature & Unit  & Symbol & Values\\
\midrule
Type of NPs & - & Type & Inorganic, Organic, Hybrid \\
Core materials of NPs & - & MAT  & Gold, Dendrimers, Liposomes, Polymeric, \newline Hydrogels, Other Organic Material, Other Inorganic Material \\
Shape of NPs    & -   & Shape  & Spherical, Rod, Plate, Others \\
Hydrodynamic diameter    & Nm   & HD  & [5, 456] \\
Zeta potential    & mV   & ZP  &   [0, 274]\\
Charge    & -   & Charge  & Positive, Negative, Neutral  \\
Targeting strategy    & -   & TS  & Passive, Active  \\
Tumor model  & -   & TM  & Allograft Heterotopic, Allograft Orthotopic, Xenograft Heterotopic, Xenograft Orthotopic  \\
Cancer type  & -   & CT  & Brain, Breast, Cervix, Colon, Liver, Lung, Ovary, Pancreas, Prostate, Skin \\
Tumor weight & g  & TW  & [0.02, 5.09] \\
Tumor size   & cm & TSiz & [0.02, 1.8] \\
Dose    & mg/kg & Dose  & [0.001, 1220] \\
Body weight & g & BW & [16,35] \\
Administrated route & - & AR & IV \\
\toprule
Prediction & Unit  & Symbol & Values\\
\midrule
Release rate constant of
tumor cells & 1/h & KTRESrelease & [0.0001, 14] \\
Maximum uptake rate
constant of tumor cells & 1/h & KTRESmax  & [[0.001, 25] \\
 Hill coefficient of tumor
cells & - & KTRESn  & [0.01, 10] \\
Time reaching half
maximum uptake rate
of tumor cells & h & KTRES50  & [0.00001, 180]  \\

\botrule
\end{tabular}
\end{table}

\subsection{Benchmarking}

The proposed multi-view AI framework was evaluated against competing models, including standard deep learning architectures~\cite{chou2023artificial}, and traditional ML methods such as Ridge Regression (RIDGE)~\cite{chou2023artificial}, RF~\cite{chou2023artificial}, Support Vector Regression (SVR)~\cite{chou2023artificial}, XGB~\cite{chou2023artificial}, and LightGBM~\cite{chou2023artificial}. 
Since the models developed in this work are for regression, the coefficient of determination ($R^2$) and root mean squared error (RMSE) are applied to evaluate model performance on the four continuous outputs, representing the four Pharmacokinetic parameters. All models are developed using Python 3.10.12, TensorFlow 2.16.1, Numpy 1.26.4, Pandas 2.1.4, and Scikit-learn 1.4.2, providing a stable and well-supported computational environment.

\subsection{Implementation Details}

Our DNN model consists of two parallel components, as shown in Fig.~\ref{fig:model}: an auxiliary MLP and a cross-attention module appended with four MLPs, one for each outcome (i.e., KTRESmax, KTRESn, KTRES50, and KTRESrelease). For all five MLPs, the number of hidden units is in the range $[64, 256]$, with dropout rates in the range $[0.1, 0.3]$. The auxiliary MLP has between 1 and 3 layers. The cross-attention mechanism and the four appended MLPs incorporate a single cross-attention layer with layer normalization and dropout rates in the range $[0.2, 0.4]$. For the four appended MLPs, an \(L_2\) regularization rate of $0.02$ is employed. The number of layers for the MLPs corresponding to KTRESmax, KTRESn, and KTRES50 lies in the range $[1, 3]$, and for the KTRESrelease pathway in the range $[1, 5]$. Batch normalization is integrated into the cross-attention mechanism to ensure stable training and accelerate convergence~\cite{ioffe2015batch}.

\subsection{Results}
\label{sec:ablation_study}

Table~\ref{tab:results_competition} compares our proposed model against competing methods in terms of $R^2$ and RMSE results for the 5-fold cross-validation and the train-validation-test experiments. In both experiments, our method shows superior results compared to competing approaches.
The enhancements produced by our method highlight the critical role of the prior information and extracted features in capturing the complex NP distribution behaviour. 
Complementing these contributions, the ensemble learning with RF, XGB, and our proposed cross-attention deep learning model further improves performance by leveraging the strengths of these diverse models, outperforming the state-of-the-art  MLP~\cite{chou2023artificial} and all baseline models in most evaluation cases, particularly for the five-fold cross-validation outcomes.

To highlight the significance of our findings, we ran the paired one-sided Wilcoxon signed-rank test, with the results summarised in Table~\ref{tab:pvalue}. This test analyses the performance results from the five folds of our cross-validation experiment, comparing the squared error of our model with that of the best-performing alternative model. At a significance level of 0.05, Table~\ref{tab:pvalue} demonstrates that the p-values for all four outcomes are below 0.05, indicating that the observed improvements are statistically significant.

\begin{table}[h]
\caption{Comparison of several competing methods and ours, in terms of $R^2$ and RMSE results both on the test set and using a 5-fold cross validation (CV).  Due to  evaluation issues identified in ~\cite{chou2023artificial}, which are detailed in the supplementary section, all models developed by the ~\cite{chou2023artificial} were re-evaluated. Best results are in \textbf{bold}, and second best are in \textit{italics}.
}\label{tab:results_competition}
\tiny
\begin{tabular*}{\textwidth}{@{\extracolsep\fill}cccccccccc}
\toprule%
\multicolumn{1}{@{}c@{}}{Model}
& \multicolumn{1}{@{}c@{}}{Measure}
& \multicolumn{2}{@{}c@{}}{KTRESmax} 
& \multicolumn{2}{@{}c@{}}{KTRES50}
& \multicolumn{2}{@{}c@{}}{KTRESn}
& \multicolumn{2}{@{}c@{}}{KTRESrelease}
\\
\cmidrule{3-4}\cmidrule{5-6}\cmidrule{7-8}\cmidrule{9-10}%
& & 5-fold CV & Test & 5-fold CV & Test  & 5-fold CV & Test &  5-fold CV & Test
\\
\midrule
RIDGE & $R^2\uparrow$ & 0.00$\pm$0.07 & 0.08 & -0.02$\pm$0.06 &  0.01 & -0.01$\pm$0.02 & -0.03 & 0.02$\pm$0.15 & \textbf{0.24} \\
& RMSE$\downarrow$ & 3.38$\pm$1.04 & 3.41 & 33.47$\pm$5.63 &  27.10 & 2.25$\pm$0.18 & 2.31 & 2.17$\pm$0.32 & \textbf{1.83} \\
SVR & $R^2\uparrow$ & -0.01$\pm$0.02 & -0.01 & -0.25$\pm$0.13 &  -0.16 & 0.09$\pm$0.09 & 0.06 & 0.04$\pm$0.02 & 0.07 \\
& RMSE$\downarrow$ & 3.44$\pm$1.10 & 3.56 & 37.43$\pm$8.23 &  29.27 & 2.14$\pm$0.23 & 2.21 & 2.15$\pm$0.21 & 2.03 \\
RF & $R^2\uparrow$& 0.06$\pm$0.10 & 0.18 & 0.09$\pm$0.15 & \textit{0.13} & \textit{0.20$\pm$0.16} & \textbf{0.13} & \textit{0.12$\pm$0.15} & 0.15 \\
& RMSE$\downarrow$& 3.27$\pm$0.96 & 3.20 & 31.62$\pm$6.58 & \textit{25.33} & \textit{1.99$\pm$0.29} & \textbf{2.13} & \textit{2.06$\pm$0.35} & 1.94 \\
XGB & $R^2\uparrow$ & \textit{0.11$\pm$0.09} & 0.25 & \textit{0.12$\pm$0.17} &  0.08 & 0.09$\pm$0.15 & 0.01 & 0.06$\pm$0.16 & 0.02 \\
& RMSE$\downarrow$& \textit{3.18$\pm$0.92} & 3.06 & \textit{31.06$\pm$6.71} &  26.07 & 2.12$\pm$0.25 & 2.28 & 2.13$\pm$0.36 & 2.09 \\
LightGBM & $R^2\uparrow$ & -0.03$\pm$0.39 & \textbf{0.33} & 0.06$\pm$0.10 &  -0.05 & 0.00$\pm$0.14 & -0.07 & 0.05$\pm$0.14 & 0.18 \\
& RMSE$\downarrow$ & 3.27$\pm$0.81 & \textbf{2.90} & 32.08$\pm$5.64 & 27.88 & 2.23$\pm$0.21 & 2.36 & 2.14$\pm$0.35 & 1.91 \\
MLP & $R^2\uparrow$ & -0.16$\pm$0.64 & 0.23 & -0.32$\pm$0.48 &  -0.01 & -0.10$\pm$0.14 & -0.07 & -0.04$\pm$0.21 & -0.06 \\
& RMSE$\downarrow$& 3.36$\pm$0.81 & 3.11 & 36.92$\pm$6.60 & 27.37 & 2.35$\pm$0.32 & 2.36 & 2.23$\pm$0.39& 2.17 \\
\midrule
Ours & $R^2\uparrow$ & \textbf{0.17$\pm$0.24} & \textit{0.18} & \textbf{0.13$\pm$0.13} &  \textbf{0.14} & \textbf{0.24$\pm$0.15} & \textbf{0.13} & \textbf{0.14$\pm$0.14} & \textit{0.19} \\
& RMSE$\downarrow$ & \textbf{3.04$\pm$1.04} & \textit{3.21} & \textbf{31.00$\pm$6.21} &  \textbf{25.18} & \textbf{1.95$\pm$0.28} & \textbf{2.13} & \textbf{2.04$\pm$0.35} & \textit{1.89} \\
\botrule
\end{tabular*}
\end{table}

\begin{table}[h]
\caption{P-values for all four outcomes, calculated using the one-sided paired Wilcoxon signed-rank test. This test analyzes the performance results from the five folds of our cross-validation experiment, comparing the squared error of our model with that of the best-performing alternative model. At a significance level of 0.05, the results indicate statistically significant improvements.}\label{tab:pvalue}%
\begin{tabular}{@{}p{2cm}p{2cm}p{2cm}p{2cm}@{}}
\toprule
 KTRESmax  & KTRES50 & KTRESn & KTRESrelease \\
\midrule
 1.5e-5  & 0.023 & 0.018 & 1.4e-5\\
\botrule
\end{tabular}
\end{table}

\begin{figure}[!h]
    \centering
    \includegraphics[width=0.9\linewidth]{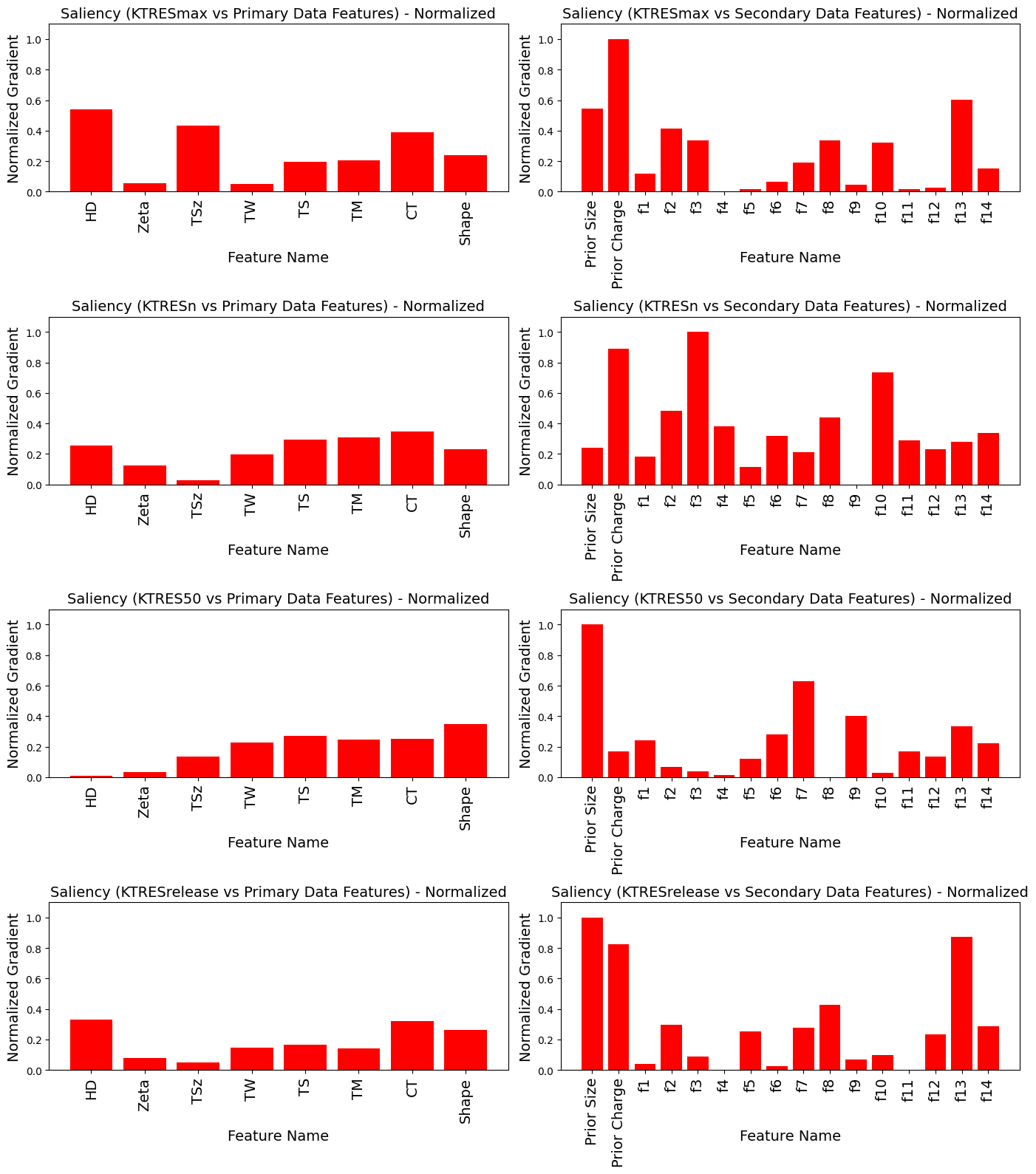}
    \caption{Saliency gradient graphs to estimate the importance of each primary (left column) and secondary (right column) features on model predictions (KTRESmax on first row, KTRESn on second row, KTRES50 on third row, KTRESrelease on fourth row) by measuring the sensitivity of outputs to changes in inputs.}
    \label{fig:saliency}
\end{figure}

We further analyse our proposed model with interpretability results to estimate the importance of each primary and secondary  features in the estimation of the outcomes KTRESmax, KTRESn, KTRES50, and KTRESrelease.
Feature importance is measured with saliency gradient graphs~\cite{simonyan2013deep}, as shown in Figure~\ref{fig:saliency}. 
Saliency gradient graphs provide a way to interpret the influence of input primary and secondary features on model predictions by measuring the sensitivity of outputs to changes in inputs. These graphs are computed by calculating the absolute gradients of the predicted outcomes (KTRESmax, KTRESn, KTRES50, and KTRESrelease) with respect to each input feature. The gradients are then averaged across multiple samples and normalized to generate feature-specific saliency scores. These scores are visualized as bar graphs, highlighting the relative importance of primary and secondary features in the model’s decision-making process.
The results show that HD (NP size) is influential across all outcomes except KTRES50, while shape, cancer type, and prior features consistently exhibit high relevance. Specifically, for KTRESmax, f13 (log of NP size) is the most salient, while Tumor Weight–derived features, such as f4 (ratio of Tumor Weight to Tumor Size) and f5 (log of Tumor Weight), are less so. For KTRESn, f3 (ratio of NP size to Tumor Weight) is most relevant; for KTRES50, f7 (ratio of Zeta to NP size) stands out; and for KTRESrelease, f13 (log of NP size) is the most significant.

\subsection{Ablation Study}\label{sec:ablation_study}

In this ablation study, we compare our final model, comprising an ensemble formed with our DNN, an RF~\cite{gavas2021nanoparticles}, and a XGB~\cite{ganaie2022ensemble}, against the following alternative configurations: 1) DNN trained with only the primary features (DNN Primary) -- in this model, the MLP does not concatenate the samples from the primary and secondary datasets, so the attention mechanism works only with the samples from the primary dataset; 2) DNN using only the secondary features (DNN Secondary) -- in this model, the MLP does not concatenate the samples from the primary and secondary datasets, so the attention mechanism works only with the samples from the secondary dataset; 3) the proposed DNN without using the ensemble model from Eq.~\eqref{eq:ensemble} (DNN); 4) the ensemble of DNN and XGB (DNN+XGB); 5) the ensemble of DNN and RF (DNN+RF); and 6) our proposed model containing the whole ensemble of DNN, XGB and RF (DNN+XGB+RF).
These experiments primarily aim to validate the effectiveness of the proposed incorporation of prior knowledge and feature engineering, and the effectiveness of the ensemble, which enhances model performance.

Table~\ref{tab:ensemble} presents the results from the ablation study, which suggests that the  incorporation of prior knowledge and extracted features (through the secondary features) provide substantial improvements in comparison with the results from the model trained with the primary or secondary features only. This confirms the significance of combining domain-specific prior knowledge (through the secondary features) with the original primary features for training our proposed model.
Table~\ref{tab:ensemble} also shows that the full ensemble configuration (``DNN+XGB+RF'') achieves the best performance across most cross-validation cases, demonstrating its robustness in combining multiple predictive models~\cite{ganaie2022ensemble}.
It is also worth noting that adding RF and XGB does not consistently translate into larger accuracy. Nonetheless, these ensemble configurations yield lower standard deviation values across cross-validation folds, indicating more stable performance than the DNN model.
Another interesting point is the observed gap between cross-validation and final test results for certain outcomes, reflecting deep learning models' large-scale data requirements for generalisation, a limitation often encountered in biomedical applications~\cite{chou2023artificial}.
In conclusion, while ensemble learning demonstrates its utility in refining predictions, the results also show that the key improvements stem from integrating prior domain knowledge and engineered features~\cite{varma2006bias, chou2023artificial}.

Table~\ref{tab:ensemble_time} shows that training is faster when using secondary features, either alone or combined with primary features, as they provide more informative signals. In contrast, relying solely on primary data requires longer training to achieve competitive performance. The multi-view approach (combining primary and secondary features) has a training time between that of using only secondary or only primary features, reflecting the added complexity and informativeness. Notably, ensembling both feature types does not significantly increase training time, maintaining practicality while improving performance.

\begin{table}[h]
\caption{Ablation study on the impact of different ensemble models. The table compares various configurations: (1) DNN trained with only primary features (DNN Primary); (2) DNN trained with only secondary features (DNN Secondary); (3) the proposed DNN model without the ensemble approach from Eq.~\eqref{eq:ensemble} (DNN); (4) the ensemble of DNN and XGB (DNN+XGB); (5) the ensemble of DNN and RF (DNN+RF); and (6) the full proposed ensemble model incorporating DNN, XGB, and RF (DNN+XGB+RF).}\label{tab:ensemble}
\tiny
\centering

\begin{adjustbox}{minipage=0.6\paperwidth,center}

\begin{tabular*}{1\textwidth}{@{\extracolsep{\fill}}cccccccccc}
\toprule%
\multicolumn{1}{@{}c@{}}{Model}
& \multicolumn{1}{@{}c@{}}{Measure}
& \multicolumn{2}{@{}c@{}}{KTRESmax} 
& \multicolumn{2}{@{}c@{}}{KTRES50}
& \multicolumn{2}{@{}c@{}}{KTRESn}
& \multicolumn{2}{@{}c@{}}{KTRESrelease}
\\
\cmidrule{3-4}\cmidrule{5-6}\cmidrule{7-8}\cmidrule{9-10}%
& & 5-fold CV & Test & 5-fold CV & Test & 5-fold CV & Test &  5-fold CV & Test\\
\midrule
DNN & $R^2\uparrow$ & -0.19$\pm$0.58 & 0.03 & -0.99$\pm$0.59 &  -0.04 & -0.33$\pm$0.29 & -0.01 & -0.34$\pm$0.33 & -0.01 \\
Primary & RMSE$\downarrow$ & 3.56$\pm$1.53 & 3.49 & 46.28$\pm$11.33 &  27.69 & 2.58$\pm$0.47 & 2.29 & 2.54$\pm$0.51 & 2.12 \\
\hline
DNN & $R^2\uparrow$ & -0.23$\pm$0.28 & -0.05 & -0.04$\pm$0.28 &  -0.08 & -0.14$\pm$0.11 & -0.03 & -1.01$\pm$1.33 & 0.03  \\
Secondary & RMSE$\downarrow$ & 3.68$\pm$1.01 & 3.63 & 33.55$\pm$7.55 &  28.20 & 2.38$\pm$0.16 & 2.31 & 3.00$\pm$1.12 & 2.07   \\
\hline
DNN & $R^2\uparrow$ & 0.16$\pm$0.44 & 0.05 & -0.05$\pm$0.10 &  0.01 & 0.24$\pm$0.14 & 0.00 & 0.01$\pm$0.13 & 0.14 \\
& RMSE$\downarrow$ & 3.00$\pm$1.63 & 3.46 & 34.24$\pm$6.99 &  27.03 & 1.94$\pm$0.24 & 2.28 & 2.18$\pm$0.32 & 1.98  \\
\hline
DNN+ & $R^2\uparrow$ & 0.15$\pm$0.32 & 0.17 & 0.10$\pm$0.13 &  0.12 & 0.22$\pm$0.16 & 0.11 & 0.13$\pm$0.14 & 0.20 \\
XGB & RMSE$\downarrow$ & 3.02$\pm$1.09 & 3.23 & 31.51$\pm$6.16 &  25.48 & 1.96$\pm$0.27 & 2.15 & 2.05$\pm$0.34 & 1.89  \\
\hline
DNN+ & $R^2\uparrow$& 0.19$\pm$0.27 & 0.13 & 0.07$\pm$0.11 &  0.13 & 0.26$\pm$0.14 & 0.10 & 0.11$\pm$0.14 & 0.19 \\
RF& RMSE$\downarrow$& 3.03$\pm$1.23 & 3.30 & 32.14$\pm$6.48 & 25.39 & 1.92$\pm$0.26 & 2.16 & 2.07$\pm$0.34 & 1.90   \\
\hline
DNN+ & $R^2\uparrow$ & 0.17$\pm$0.24 & 0.18 & 0.13$\pm$0.13 &  0.14 & 0.24$\pm$0.15 & 0.13 & 0.14$\pm$0.14 & 0.19  \\
XGB+RF& RMSE$\downarrow$& 3.04$\pm$1.04 & 3.21 & 31.00$\pm$6.21 & 25.18 & 1.95$\pm$0.28 & 2.13 & 2.04$\pm$0.35 & 1.89 \\
\botrule
\end{tabular*}%

\end{adjustbox}

\end{table}

\begin{table}[h]
\caption{Training and testing time of the models presented in Tab.~\ref{tab:ensemble}.}\label{tab:ensemble_time}
\tiny
\centering

\begin{adjustbox}{minipage=0.6\paperwidth,center}

\begin{tabular*}{0.75\textwidth}{@{\extracolsep{\fill}}ccc}
\toprule%
\multicolumn{1}{@{}c@{}}{Model}
& \multicolumn{2}{@{}c@{}}{Time Taken}
\\
\cmidrule{2-3}%
& Training & Testing \\
\midrule
DNN Primary & 14 hrs 25 min & 1 sec\\
DNN Secondary & 10 hrs 11 min & 0.5 sec \\
DNN & 12 hrs 24 min & 0.5 sec \\
DNN+XGB & 12 hrs 26 min &  1 sec \\
DNN+ RF& 12 hrs 24 min &  1 sec \\
DNN+XGB+RF & 12 hrs 26 min &  1 sec\\
\botrule
\end{tabular*}%

\end{adjustbox}

\end{table}

\subsection{Discussion}

This AI-driven approach offers a powerful tool for streamlining NP design and formulation by identifying key physicochemical parameters that influence biodistribution. Our model highlights critical NP properties, such as surface charge and hydrodynamic diameter, that significantly impact organ-level accumulation \cite{sodipo2024advances,yu2024state}. Leveraging these insights enables the refinement of NP formulations to achieve targeted biodistribution patterns while minimizing reliance on time-consuming and costly in vivo screening.  

By integrating AI-powered pharmacokinetic modelling into the research pipeline, our study demonstrates its potential as a pre-screening tool to complement experimental studies. This approach can reduce the number of required biological assays by narrowing down formulation candidates before in vivo testing, allowing researchers to focus on the most promising NPs. As a result, the optimisation process is accelerated, improving translational efficiency and expediting the development of clinically relevant nanomedicines.  

Beyond enhancing formulation efficiency, our study underscores the value of integrating AI with experimental workflows. Rather than replacing traditional experimental methods, AI enhances predictive power by incorporating real-world biological insights. Future efforts should focus on validating AI predictions across diverse NP classes to further strengthen the synergy between AI-driven modelling and experimental nanomedicine.

\section{Methods}
\label{sec:methods}

In this section, we present our method, emphasising our two main technical contributions: (1) the introduction of new prior knowledge and extracted features to improve the training effectiveness of NP delivery prediction using small datasets; and (2) the development of an advanced deep learning model capable of leveraging the original raw data from the dataset together with the prior knowledge and the extracted features from (1).

\subsection{Dataset, Prior Knowledge, and Feature Extraction}

To predict NP pharmacokinetics and biodistribution across target organs such as the liver, spleen, lungs, and kidneys, we rely on two datasets to inform our model. 

The primary dataset, denoted as $\mathcal{D} = \{(x_i, y_i)\}_{i=1}^{N}$, contains the original feature vector \( x_i \in \mathcal{X} \subseteq \mathbb{R}^d \) for each sample \( i \), where \( d \) is the number of features in \( x \). The target vector \( y_i \in \mathcal{Y} \subseteq \mathbb{R}^4 \) includes four continuous pharmacokinetic parameters—KTRESmax, KTRESn, KTRES50, and KTRESrelease—describing NP biodistribution across the organs: maximum uptake rate constant of tumor cells, Hill coefficient of tumor cells, time reaching half maximum uptake rate of tumor cells, and release rate constant of tumor cells.
~\cite{peer2020nanocarriers,brigger2012nanoparticles,davis2008nanoparticle,chou2023artificial}.

The secondary dataset \( \mathcal{D}_{\text{P+FE}} = \{\tilde{x}_i\}_{i=1}^{N} \) contains both the prior knowledge and the extracted features to provide additional insights into NP-organ interactions to aid the prediction of NP biodistribution across the organs. 
Each transformed feature vector \( \tilde{x}_i \in \mathcal{X}_{\text{P+FE}} \subseteq \mathbb{R}^{16} \) is derived by applying the transformation \( f_{\text{P+FE}}: \mathcal{X} \to \mathcal{X}_{\text{P+FE}} \) to \( x \), as:
\[
f_{\text{P+FE}}(x) = \left(f_{\text{size}}(x), f_{\text{charge}}(x), f_{1}(x), \dots, f_{14}(x)\right)
\]
where \( f_{\text{size}}:\mathcal{X} \to \mathbb{R} \) and \( f_{\text{charge}}:\mathcal{X} \to \mathbb{R} \) represent the prior knowledge, and functions \( f_{1}:\mathcal{X} \to \mathbb{R} \) through \( f_{14}:\mathcal{X} \to \mathbb{R} \) represent the  extracted features~\cite{yeh2018building,heaton2016empirical,hammer2020relationship}. 

\subsection{Incorporating Prior Knowledge}

To model the prior knowledge on the NP absorption in the liver, spleen, lungs, and kidneys, we use functions \( f_{\text{size}}(.) \) and \( f_{\text{charge}}(.) \), which evaluate NP properties against organ-specific criteria, capturing essential size- and charge-related bio-distribution patterns respectively. This approach allows the model to approximate each organ’s unique absorption behaviour by encoding relevant, simplified interactions.

The size-based function reflects how particle size influences NP bio-distribution across different organs, with:
\begin{equation}
f_{\text{size}}(x) = \sum_{j \in \mathcal{O}} s_j(x), \text{ where } s_j(x) = 
\begin{cases} 
1, & \parbox[t]{\textwidth}{if NP size aligns with size-related \newline  absorption characteristics for organ \( j \)} \\
0, & \text{otherwise}
\end{cases}
\label{eq:f_size}
\end{equation}
where \( \mathcal{O} = \{ \text{kidney, spleen, liver, lung} \} \). Small NPs (typically below six nanometers) are more likely to be filtered out by the kidneys, whereas larger NPs can be captured by macrophages in the spleen. The liver’s uptake spans a broad size range due to contributions from both hepatocytes and Kupffer cells, while the lungs show preferential uptake for smaller particles reaching the alveolar region~\cite{choi2007renal, sonavane2008biodistribution, he2024understanding, lu2014right}.

The charge-based function approximates the influence of NP surface charge on organ-specific absorption patterns, with
\begin{equation}
f_{\text{charge}}(x) = \sum_{j \in \mathcal{O}} c_j(x), \text{ where } c_j(x) = 
\begin{cases} 
1, & \parbox[t]{\textwidth}{if NP charge aligns with charge-related \newline  absorption characteristics for organ \( j \)} \\
0, & \text{otherwise}
\end{cases}
\end{equation}
where \( \mathcal{O} \) is defined in~\eqref{eq:f_size}. For instance, positively charged NPs often interact with liver hepatocytes, while neutral or negatively charged NPs are more likely to interact with Kupffer cells or immune cells in the spleen. In the lungs, positively charged particles can interact with pulmonary surfactants, while neutral NPs have greater potential for renal clearance~\cite{he2010effects, mousseau2018role, hammer2020relationship}. This framework enables the model to capture essential organ-specific responses to size and charge, preserving key biodistribution patterns with relatively simple interactions.

\subsection{Incorporating Extracted Features}

The features extracted from the primary data \(x \in \mathcal{X}\) are defined as:
\begin{equation}
\begin{split}
f_1(x) &= \phi_{\text{ratio}}(x({\text{NP Size}}),\ x({\text{Tumor Size}})) \\[5pt]
f_2(x) &= \phi_{\text{ratio}}(x({\text{NP Size}}),\ x({\text{Zeta}})) \\[5pt]
f_3(x) &= \phi_{\text{ratio}}(x({\text{NP Size}}),\ x({\text{Tumor Weight}})) \\[5pt]
f_4(x) &= \phi_{\text{ratio}}(x_({\text{Tumor Weight}}),\ x({\text{Tumor Size}})) \\[5pt]
f_5(x) &= \phi_{\text{log}}(x({\text{Tumor Weight}})) \\[5pt]
f_6(x) &= \phi_{\text{polynomial}}(x({\text{Tumor Weight}})) \\[5pt]
f_7(x) &= \phi_{\text{ratio}}(x({\text{Zeta}}), \ x({\text{NP Size}})) \\[5pt]
f_8(x) &= \phi_{\text{interaction}}(x({\text{NP Size}}),\ x({\text{Zeta}}),\ x({\text{Charge}})) \\[5pt]
f_9(x) &= \phi_{\text{interaction}}(x({\text{Zeta}}),\ x({\text{Shape}}),\ x({\text{Charge}}))) \\[5pt]
f_{10}(x) &= \phi_{\text{polynomial}}(x({\text{Tumor Size}})) \\[5pt]
f_{11}(x) &= \phi_{\text{polynomial}}(x({\text{Zeta}})) \\[5pt]
f_{12}(x) &= \phi_{\text{log}}(x({\text{Tumor Size}})) \\[5pt]
f_{13}(x) &= \phi_{\text{log}}(x({\text{NP Size}})) \\[5pt]
f_{14}(x) &= \phi_{\text{interaction}}(x({\text{Tumor Size}}),\ x({\text{NP Size}}))),
\end{split}
\label{eq:extracted_features}
\end{equation}
where \(\phi_{\text{ratio}}\) divides the first argument by the second argument; 
\(\phi_{\text{log}}\) applies a logarithm to the input; 
\(\phi_{\text{polynomial}}\) squares the input; 
and \(\phi_{\text{interaction}}\) multiplies all input arguments.
The functions $f_1(.)$ to $f_{14}(.)$ in Eq.~\eqref{eq:extracted_features} are engineered to encapsulate crucial relationships between NP properties and tumor characteristics by applying ratio, interaction, polynomial, and logarithmic transformations. These features are designed to allow the model to effectively capture complex biological behaviors influencing NP biodistribution. For instance, ratios between NP size and tumor size or weight can reflect how particle dimensions affect accumulation within tumors~\cite{brigger2012nanoparticles,sonavane2008biodistribution}.Interaction terms involving NP size, zeta potential (a measure of the surface charge of NPs), and charge consider how combined physical and chemical properties impact cellular uptake and distribution.~\cite{he2010effects,mousseau2018role}. Polynomial transformations emphasize the influence of larger values, such as tumor size or NP surface charge, on biodistribution patterns~\cite{zhang2016nanoparticle,aghebati2020nanoparticles}. Logarithmic transformations help in normalizing skewed data distributions, improving model stability and prediction accuracy~\cite{wang2013metabolism}. By integrating these features, the model is consistent with advanced methodologies in nanomedicine and pharmacokinetic modeling, enhancing its predictive power for NP behavior in complex biological systems~\cite{scarpa2020tuning,he2024understanding}.

\subsection{Model Architecture and Cross-Attention Mechanism}

We introduce three different model architectures to leverage the data from \(\mathcal{D}\) and the prior and features from $\mathcal{D}_{\text{P+FE}}$: the cross-attention based deep learning model; the self-attention deep learning model; and the multi-layer perceptron (MLP). The cross-attention model focuses on estimating the interactions between two separate input channels: one channel containing the raw data $x$ from $\mathcal{D}$, and another channel with the prior and features $\tilde{x}$ from $\mathcal{D}_{\text{P+FE}}$.
In contrast, the self-attention model and MLP models have a single input channel formed from the data $x$ from $\mathcal{D}$ and $\tilde{x}$ from $\mathcal{D}_{\text{P+FE}}$ concatenated. 
These three models aim to predict the four pharmacokinetic parameters \( y \in \mathcal{Y} \).

\begin{figure}[h!]
    \centering
    \includegraphics[width=\linewidth]{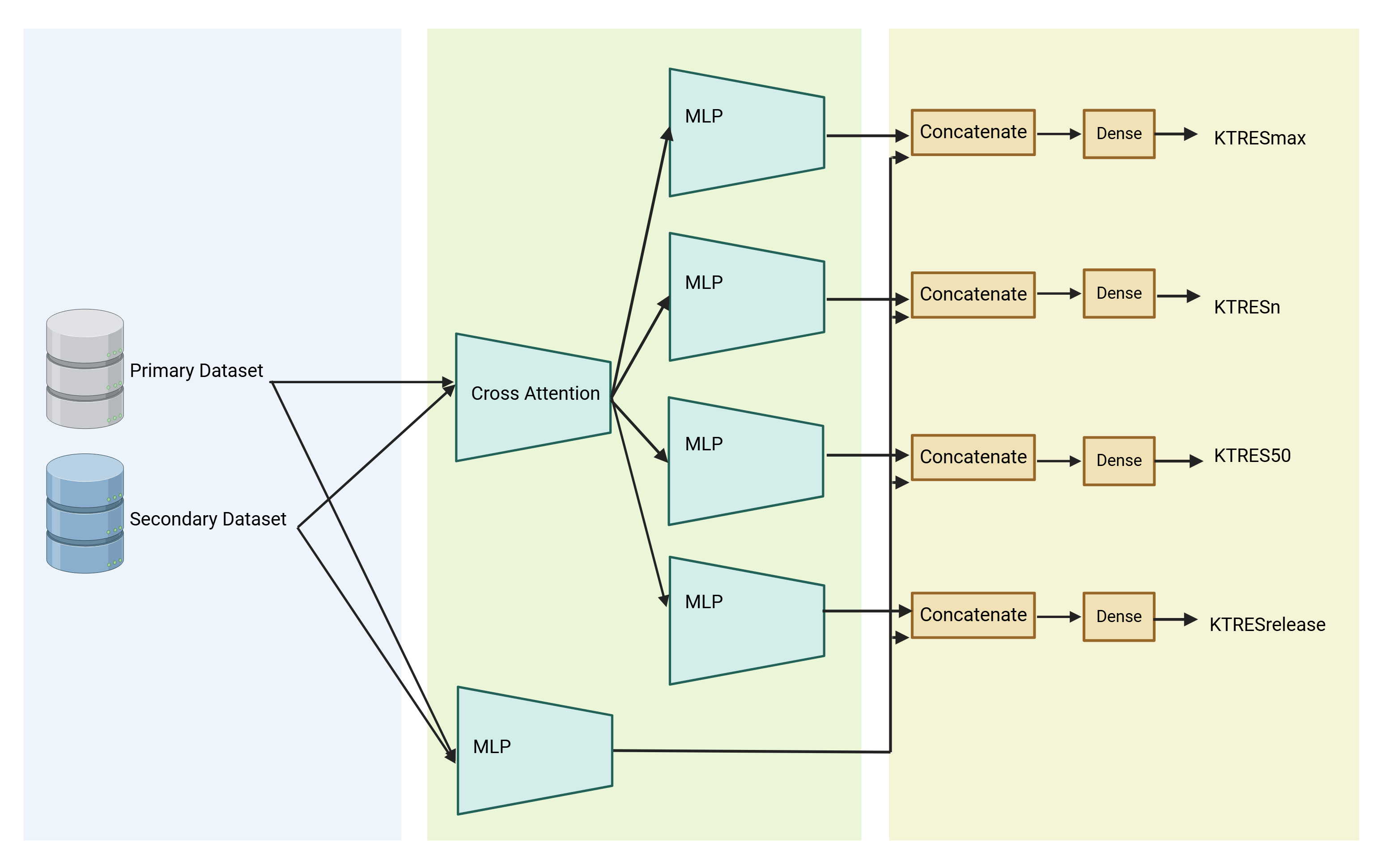}
    \caption{Overview of the proposed multi-task multi-view model architecture, illustrating the cross-attention and MLP branches that are then concatenated to produce a final prediction. In this figure, the primary dataset represents \( \mathcal{D} \), and the secondary dataset \( \mathcal{D}_{\text{P+FE}} \). Created in BioRender. Khakpour, A. (2025) https://BioRender.com/r59r002
}
    \label{fig:model}
\end{figure}

The cross attention model, shown in Fig.~\ref{fig:model}, represents a multi-view model architecture that combines the two input channels using a cross-attention mechanism, which allows a more nuanced integration compared to simpler approaches like input data concatenation~\cite{vaswani2017attention}. Specifically, the cross-attention mechanism derives the key matrix  \( K \) from \( x \in \mathcal{D} \), and the query matrix \( Q \) and value matrix \( V \)  from \( \tilde{x} \in \mathcal{D}_{\text{P+FE}} \). This configuration enables the model to effectively integrate complementary information from both data sources, enhancing the quality of predictions in pharmacokinetic modeling~\cite{wang2013metabolism}.
The final predicted output \( \hat{y} \in \mathcal{Y} \) is computed by the model:
\begin{equation}
\hat{y} = f_{\theta}( \text{CrossAttention}(\mathsf{Dense}_x(x), \mathsf{Dense}_{\tilde{x}}(\tilde{x})) )
\label{eq:model}
\end{equation}
where $\mathsf{Dense}_x(.)$ and $\mathsf{Dense}_{\tilde{x}}(.)$ represent dense layers that transform the input data into a common dimensional feature space, and \( f_{\theta} \) represents the deep learning model parametrized by \( \theta \in \Theta \). The cross-attention function~\cite{vaswani2017attention} jointly analyses features from both views, supporting an integrated prediction of the target pharmacokinetic parameters.
This design is motivated by the need for the model to leverage distinct aspects of the primary dataset \( \mathcal{D} \) and the secondary dataset \( \mathcal{D}_{\text{P+FE}} \), where the cross-attention mechanism allows the model to focus on critical features in both datasets simultaneously, learning how NP properties interact with tumor characteristics. 
Our model includes a parallel branch to the cross attention, containing an MLP model that also processes concatenated data from the primary dataset \( \mathcal{D} \) and the secondary dataset \( \mathcal{D}_{\text{P+FE}} \), capturing additional patterns without relying on cross-attention. 
This parallel structure is inspired by  AlphaFold, which integrates multiple views of the data to make more comprehensive predictions~\cite{jumper2021highly}.
The outputs from both the cross-attention and the MLP branches are then concatenated, and a final MLP is employed to produce the final prediction. 
This final MLP allows the model to produce a robust prediction incorporating both feature-specific and cross-view insights, significantly enhancing the prediction accuracy for selected outputs.

We combine our cross-attention-based deep learning model with traditional machine learning models, such as RF~\cite{gavas2021nanoparticles} and XGB~\cite{ganaie2022ensemble}, to leverage their complementary strengths and produce more robust and accurate predictions. In particular, these traditional models are particularly effective at handling tabular data and small datasets, while our cross-attention-based deep learning model excels in capturing complex relationships between the original and engineered features.

The final prediction is calculated with a weighted average of outputs from the deep learning model and the traditional machine learning models, as:
\begin{equation}
    \hat{y}_{\text{ensemble}} = \omega_0 \times \hat{y} + \sum_{k=1}^{K} \omega_{k} \times  \hat{y}_k,
    \label{eq:ensemble}
\end{equation}
where $\hat{y}$ is computed from (4), $\hat{y}_k$ represents the predictions of the $k$-th traditional model, and the weights $\{\omega_k\}_{k=0}^{K}$ are estimated based on validation performance across multiple metrics~\cite{ganaie2022ensemble, aghebati2020nanoparticles}. By integrating these diverse approaches, our method combines the strengths of both deep learning and traditional methods, leading to more accurate predictions.

\subsection{Optimization}

We estimate the parameters of the model defined in~\eqref{eq:model} by minimizing the Mean Squared Error (MSE) loss with \(L_2\) regularization applied to different layers of the model, as follows:
\begin{equation}
\theta^* = \arg\min_{\theta \in \Theta} \frac{1}{N} \sum_{i=1}^{N} \ell_{SE}(y_i,\hat{y}_i) + \lambda \sum_{l=1}^{L} \|\theta_l\|_2^2,
\label{eq:optimization}
\end{equation}
where \( y_i \) is the target, \( \hat{y}_i \) is the value predicted by the model \( f_{\theta}(.) \) in~\eqref{eq:model} for sample \( i \)~\cite{sekeroglu2022comparative}, and \( \ell_{SE}(y_i,\hat{y}_i) = \| y_i - \hat{y}_i\|_2^2 \) represents the squared error loss. The term \( \sum_{l=1}^{L} \|\theta_l\|_2^2 \) represents layer-wise \(L_2\) regularization, where \( \theta_l \) are the parameters of the \( l \)-th layer, and \( L \) is the total number of layers. The hyperparameter \( \lambda \) controls the regularization strength, preventing overfitting by penalizing large weights.  

When working with the ensemble model from Eq.~\eqref{eq:ensemble}, we use the same optimization from Eq.~\eqref{eq:optimization}, where we estimate not only the model parameters \( \theta \), but also the machine learning models' weights \( \{ \omega_k \}_{k=0}^{K} \).  

Combining prior knowledge, extracted features, and the information from the original samples using a cross-attention mechanism, our methodology offers an accurate and robust approach for predicting NP bio-distribution and pharmacokinetics.


\section{Conclusion}\label{sec:conclusion}

This study presents a novel multi-view AI framework that integrates prior knowledge with deep learning to enhance NP pharmacokinetic predictions. By leveraging cross-attention mechanisms and ensemble learning, the proposed model demonstrates significant improvements over conventional approaches, offering a scalable solution for precision nanomedicine and AI-driven drug delivery optimization. Importantly, our results illustrate how AI can assist in refining NP formulations, optimising drug delivery strategies, and reducing the reliance on exhaustive biological screening. Future research will focus on expanding the dataset, integrating multi-modal data sources, and further validating the model across diverse NP formulations to enhance its clinical applicability.

\backmatter

\bmhead{Acknowledgements}

Support from the Australian Research Council and University of Surrey is acknowledged.


\bibliography{sn-bibliography}

\section{Appendix}
\subsection{Supplementary Information}

While analyzing the implementation of the state-of-the-art model from Chou et al.~\cite{chou2023artificial}, we identified two key issues in their evaluation process for the deep neural network (DNN) model during cross-validation.

\begin{enumerate}
    \item \textbf{Untrained Models in Cross-Validation:} The DNN model used for evaluation in each fold of cross-validation was not initialized as a new, untrained model. This could lead to model performance being influenced by residual learning from previous folds, thereby compromising the validity of the cross-validation results.
    
    \item \textbf{Absence of Validation Set:} The state-of-the-art implementation did not include a separate validation set during the training of the DNN. This omission likely affected the ability of the model to generalize effectively, as hyperparameter tuning and model checkpoints were directly influenced by the test fold.
\end{enumerate}

To address these issues, we updated the code to ensure:
\begin{itemize}
    \item A fresh, untrained instance of the DNN model was used for each fold of cross-validation.
    \item A separate validation set was included for the DNN during training in each fold, allowing for effective model monitoring and parameter tuning.
\end{itemize}

By implementing these corrections, we ensured a rigorous and fair evaluation of our proposed model and baseline methods. The updated methodology provided a more accurate comparison between our approach and the state-of-the-art methods.

\end{document}